\documentclass[runningheads]{llncs}
\usepackage{graphicx}
\usepackage[english]{babel}
\usepackage[utf8]{inputenc}
\usepackage{algorithm}
\usepackage{algcompatible}
\usepackage{wrapfig}
\usepackage{lipsum}
\usepackage{eucal}
\usepackage{dsfont}
\usepackage{color}
\usepackage{multirow}
\usepackage{array}
\usepackage{adjustbox}
\usepackage{caption}

\newcolumntype{R}[2]{%
    >{\adjustbox{angle=#1,lap=\width-(#2)}\bgroup}%
    l%
    <{\egroup}%
}

\begin{document}
%
\title{Data Augmentation with Manifold Exploring Geometric Transformations for Increased Performance and Robustness}
%
\titlerunning{Manifold Exploring Data Augmentation}
%
\author{Magdalini Paschali\inst{1} \and
Walter Simson\inst{1} \and
Abhijit Guha Roy\inst{2,1} \and
Muhammad Ferjad Naeem\inst{1} \and
R{\"u}diger G{\"o}bl\inst{1} \and
Christian Wachinger\inst{2}\and Nassir Navab\inst{1,3}}



\authorrunning{Paschali et al.}

 \institute{Computer Aided Medical Procedures, Technische Universität München, Germany \\ \email{magda.paschali@tum.de} \and
 Department of Child and Adolescent Psychiatry, Psychosomatic and Psychotherapy, Ludwig-Maximilian-University, Munich, Germany \and Computer Aided Medical Procedures, Johns Hopkins University, USA}
\maketitle              
\begin{abstract}
In this paper we propose a novel augmentation technique that improves not only the performance of deep neural networks on clean test data, but also significantly increases their robustness to random transformations, both affine and projective. Inspired by ManiFool, the augmentation is performed by a line-search manifold-exploration method that learns affine geometric transformations that lead to the misclassification on an image, while ensuring that it remains on the same manifold as the training data. 

This augmentation method populates any training dataset with images that lie on the border of the manifolds between two-classes and maximizes the variance the network is exposed to during training. Our method was thoroughly evaluated on the challenging tasks of fine-grained skin lesion classification from limited data, and breast tumor classification of mammograms. Compared with traditional augmentation methods, and with images synthesized by Generative Adversarial Networks our method not only achieves state-of-the-art performance but also significantly improves the network's robustness.
\end{abstract}
\section{Introduction}

\begin{figure}[htb]
\centering
  \includegraphics[width=0.95\linewidth]{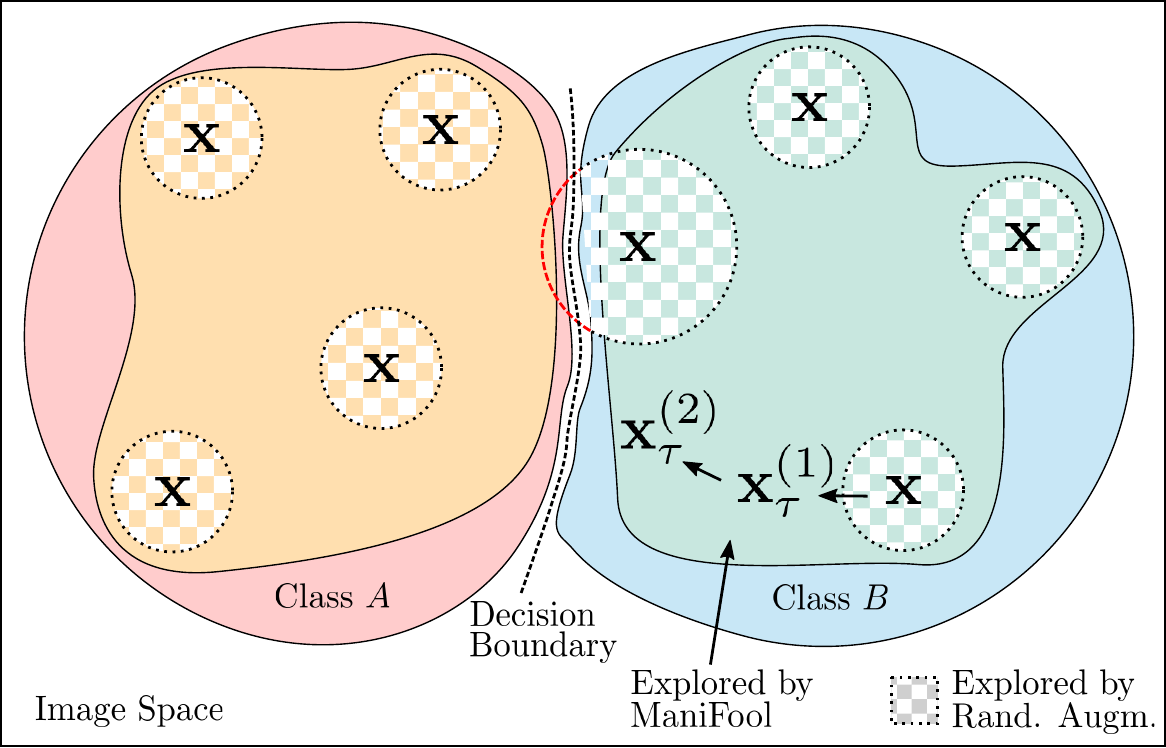}
  \caption{\small{\textbf{Schematic representation of proposed augmentation}: The proposed augmentation scheme based on ManiFool explores the present classes towards the decision boundaries, thus adding more relevant training samples $\textbf{x}_\tau^{(i)}$ than random augmentation (checkerboard pattern) which explores the space around the original training samples $\textbf{x}$ locally. Additionally, it is ensured that samples from ManiFool Augmentation originate from the ground truth class.}} 
  \label{fig:overview}
\end{figure}

Recently, medical imaging tasks such as classification, segmentation and registration have been successfully carried out with state-of-the-art performance by deep learning models, which have found their way into a plethora of Computer Assisted Diagnosis and Intervention (CAD/I) Systems which aid physicians.
However, medical imaging datasets utilized to train such models are often characterized by large class variability, severe class imbalance, outliers, inter-observer variability, ambiguity and most prominently limited data.
The aforementioned problems hinder the training of neural networks and lead to sub-optimal and overfit solutions.
Moreover, deep learning models deployed by physicians in a CAD/I system must be thoroughly evaluated, with respect to not only their generalizability, i.e. performance on data originating from a given test set, but also their behavior on data corrupted by noise, unknown transformations and outliers, which can be described by the term robustness.
Data augmentation describes the act of increasing the size and variance of a given dataset to train a machine learning model, in order to achieve better generalizability and capture a better understanding of the underlying distribution of the training data.
The manifold of a class learned by a classifier can be perceived as the space that represents the distribution of the training data.


In this work our contribution is two-fold: We propose a novel data augmentation technique, utilizing an exhaustive manifold-exploration method that increases the performance of a deep learning model on the provided test set, and significantly improves its robustness to random geometric transformations. Furthermore, we provide quantitative measures to assess a classifier's robustness. Such a measure provides a significant step towards a thorough evaluation of machine learning models; a highly valuable step towards the safe and successful deployment of trained models by physicians in real-world scenarios involving patient diagnosis and treatment.

ManiFool Augmentation is performed by populating the training dataset for a given task with samples transformed with optimized affine geometric transformations.
The method is outlined in Fig.~\ref{fig:overview}, where it is contrasted with traditional data augmentation performed with random transformations.
The algorithm utilized to craft samples leveraged for data augmentation is inspired by ManiFool~\cite{manifool} (discussed in Section~\ref{sec:manifool_explanation}) and the intuition behind it is rather simple: Move an image via affine geometric transformations iteratively towards a classifier's decision boundary by following the direction that maximizes the gradient. 
After every step, project the calculated movement back onto the original training manifold of the class of the image being transformed.
This process is repeated iteratively until either a transformation is found that causes the network to misclassify the transformed sample or a pre-defined maximum amount of steps is reached.
In case of misclassification, we have crossed the decision boundary and stepped on the manifold of another class. We then backtrack to the manifold of the original class and use this calculated transformated for data augmentation during training.

Contrary to traditional augmentation methods with random transformations, ManiFool Augmentation ensures that the space explored by the network during training is not limited to the local vicinity of a training sample. Instead, augmentations are found globally up to the edges of each class-manifold for the whole training set as can be seen in Fig.~\ref{fig:overview}.
An effective augmentation technique should be able to ensure that the samples leveraged to increase the population of the training dataset originate from the same manifold as the original data. 
Augmenting the training dataset with samples from a different distribution would not necessarily facilitate the model with learning a better embedding for each of the classes, but would rather encourage it, to map the same class to two different sub-spaces, one for each training manifold.

Exhaustive experimentation on two challenging medical datasets showcases that the proposed augmentation technique does not only increase the robustness of a model to geometric transformations, but it also significantly improves its performance on the original test data.
This is additionally highlighted by cross-dataset testing, where networks trained with ManiFool Augmentation were able to better capture the underlying distribution of the training data.
\paragraph{\textnormal{\textbf{Related Work}}}\label{related_work}
Many have taken steps in addressing the problem of limited data in deep learning applications in order to improve model accuracy without carrying the burden of costly data acquisition. 
Approaches range from elastic transformations~\cite{when_to_wrap}, noise generation in a learned features space~\cite{augmentation_feature_space}, to 
repeat, rotate and infill approaches whereby a known sample is scaled and rotated in a grid pattern, and background consistency is ensured~\cite{rotation_color_consistency}.
Fawzi et. al. proposed an algorithm for augmentation which can be integrated into the process of stochastic gradient decent and seeks an augmented sample with the greatest loss within a constrained exploration space or "trust region"~\cite{adaptive_fawzi}.

Data augmentation has also been extensively formulated as a learning task.~\cite{dagan_karras} show significant improvement in accuracy of hand-written-digit classification with a method deploying DAGAN. 
AutoAugment, formulates the augmentation task as a discrete search problem in which the search algorithm itself is based on a reinforcement learning approach that strives to "learn" how to maximize the total classification accuracy via augmentation~\cite{reinforcement_learning_augmentation}.

Specifically in the field of medical deep learning applications, creative augmentation approaches are necessary to combat the extreme lack of annotated data.~\cite{gan_rueckert} employed generated augmented samples and annotations via GANs to improve CT brain segmentation under severe lack of training data.~\cite{gan_isbi} reported improved accuracy for liver segmentation by employing DCGANs for data augmentation.


\section{Method}\label{methodology}\label{sec:manifool_explanation}
ManiFool~\cite{manifool} is an iterative algorithm that can be applied to any differentiable classifier $f$. 
In this Section we will discuss the mathematical operations that generate a geometrically transformed example leveraged for data augmentation.
\paragraph{\textnormal{\textbf{Movement Direction}}}
For an image $\mathbf{x}$ with ground truth label $l$ and a binary classifier $f$ an iterative process of $i$ steps is initialized and the original image can be defined as $\mathbf{x}^{(0)}$. Initially, ManiFool finds the movement direction $\mathbf{u}$ towards the decision boundary of $f$, by following the opposite of the gradient, $-\nabla{f(\mathbf{x})}$.
The gradient at the step $i$ for the image $\mathbf{x^{(i)}}$ is the projection of  $\nabla{f(\mathbf{x^{(i)}})}$ onto the tangent space and can be calculated utilizing the pseudoinverse operation:

\begin{equation}
\mathbf{u} = - \mathbf{J_{{x}^{(i)}}^+}\nabla f( 
\mathbf{x}^{(i)}) = -(\mathbf{J_{x^{(i)}}^T} \mathbf{J_{x^{(i)}}})^{-1}\mathbf{J_{{x}^{(i)}}^T} \nabla f(
\mathbf{x}^{(i)}).
\end{equation}
$\mathbf{J_{x^{(i)}}}$ is the Jacobian matrix and the calculated $\mathbf{u}$ is the direction towards the decision boundary for step $i$.

To improve the accuracy and convergence speed during the calculation of $\mathbf{u}$ a manifold optimization technique similar to~\cite{manifold_optimization} has been adopted:
\begin{equation}
\mathbf{u}^{(i)} = -\lambda_i \frac{\mathbf{J_{{x}^{(i)}}^+}\nabla f( 
\mathbf{x}^{(i)})}{||\mathbf{J_{{x}^{(i)}}^+}\nabla f( 
\mathbf{x}^{(i)})||} + \gamma \mathbf{u}^{(i-1)},
\end{equation}
where $\lambda_i$ is the calculated step size of the iteration and $\gamma$ is a constant momentum.

\paragraph{\textnormal{\textbf{Mapping onto the original manifold}}}
After the movement direction $\mathbf{u}$ is calculated it is mapped back onto the manifold $\mathcal{M}$ of the ground truth class. Following~\cite{manifool}, this mapping is performed using retraction $R_{\mathbf{x^{(i)}}}(\mathbf{u}) = \mathbf{x_{\tau_i}^{(i)}}$, where $\tau_i$ is the affine transformation calculated as:
\begin{equation}
\tau_i = \exp \Bigg(\sum_j u_j Gj\Bigg).
\end{equation}
$G_j$ are the basis vectors of the Lie Group $\mathcal{T}$ of the calculated affine geometric transformation.
There are two conditions for the termination of the algorithm, namely the misclassification of the calculated transformed image by the model or reaching the maximum number of allowed iterations $i_{\max}$. After $i_{\max}$ steps the accumulative affine transformations applied to $\mathbf{x^{(0)}}$ to generate the ManiFool sample are given by:
\begin{equation}
\hat{\tau} = \tau_0 \circ \tau_1 \circ \dots \tau_{I_{\max}}.
\end{equation}
\vspace{-0.2cm}
\paragraph{\textnormal{\textbf{Multi-class Classifiers}}}
The extension of the method from binary to multi-class classifiers is straightforward: We generate a ManiFool sample for each of the remaining classes, starting from the ground truth and based on the geodesic distance $l$ of the transformed to the original image we leverage the sample with the smallest transformation $\tau_{l_{\min}}$. The class with the smallest geodesic distance between the transformations can be found by:
\begin{equation}
l_{\min} = \arg\min_{l \neq l_x} \tilde{d}_{\mathbf{x}^{(0)}}(e,\tau_l).
\end{equation}
In the following subsections we discuss how the distance $\tilde{d}_{\mathbf{x}^{(0)}}$ is calculated and the significant role it plays as a measure of robustness for neural networks.

\subsection{Invariance to Geometric Transformations}
\paragraph{\textnormal{\textbf{Geodesic Distance Between Transformations}}}
The geodesic distance $d_{\mathbf{x}^{(i)}}$ between two transformations $\tau_1$ and $\tau_2$ is the length $L$ of the shortest curve $\gamma$ between $\tau_1$ and $\tau_2$. However, since the metric space of the manifold of the training data is unknown we have to acquire a metric in the Riemannian space by mapping the Lie group $\mathcal{T}$ to the differentiable image manifold of $\mathbf{x}^{(i)}_{\tau_1}$ and $\mathbf{x}^{(i)}_{\tau_2}$, which inherits the Riemannian metric from $L_2$~\cite{pattern_transfrormation_manifolds,differential_geometry}. After this mapping, the geodesic distance between $\tau_1$ and $\tau_2$ is equal to the shortest path connecting $\mathbf{x}^{(i)}_{\tau_1}$ and $\mathbf{x}^{(i)}_{\tau_2}$, formulated as:
\begin{equation}
d_{\mathbf{x}^{(i)}}(\tau_1,\tau_2) = \min L(\gamma).
\end{equation}
\paragraph{\textnormal{\textbf{Geodesic Distance Between Original and ManiFool Samples}}}
Having explained how to calculate the distance between two transformations and two transformed images, we can now show how to measure the geodesic distance between the original samples of our training dataset and the ones generated with ManiFool.
The initial untransformed image $\mathbf{x}^{(0)}$ can be considered the initial point of the aforementioned $\gamma$ curve if we define its transformation $e$ as the identity one. Thus, the distance between the original sample $\mathbf{x}^{(0)}_{e}$ and $\mathbf{x}^{(i_{\max})}_{{\tau_{i_{\max}}}}$, can be calculated by the distance between the identity transformation $e$ and the final aggregated one $\tau_{i_{\max}}$:
\begin{equation}
\tilde{d}_{\mathbf{x}^{(i)}}(e,\tau_i) = \frac{d_{\mathbf{x}^{(i)}}(e,\tau)}{||\mathbf{x}^{(i)}||_{L^2}}.
\end{equation}
Normalization of the distance by the norm of the image is crucial, to ensure generalizability of the distance measure.

\paragraph{\textnormal{\textbf{Robustness to Geometric Transformations}}}\label{ssec:robust_to_geometric_transformation}
Since every computed ManiFool example originates from the edge of a class manifold, measuring the aforementioned distance $\tilde{d}_{\mathbf{x}^{(i_{\max})}}$ between an original image and its respective transformed sample can act as a measure for the robustness of a classifier. Specifically networks that have learned a high-dimensional embedding space characterized by high class compactness and maximized distance between decision boundaries will require a larger average $\tilde{d}$ to transform a class from one class to another. In this work we compute the average distance $\tilde{\rho}_{\tau}$ of all the ManiFool samples as:
\begin{equation}
    \tilde{\rho}_{\tau}(f) = \frac{1}{m} \sum_{j=1}^m\tilde{d}_{\mathbf{x}^{(i)}_j}(e,\tilde{\tau}),
    \label{eq:robustness_measure}
\end{equation} 
where $m$ is the number of crafted samples. $\tilde{\rho}_{\tau}$ acts as a quantitative measure of robustness of a neural network to geometric transformations, that can be used to compare the robustness of different deep model architectures or models trained with different augmentation techniques.

Another measure to quantify the robustness of classifier $f$ is $r_{\tau}$, given by Equation~\ref{eq:random_robustness}. $r_{\tau}$ assesses a model's performance when it's evaluated on randomly transformed images. Specifically, for a range of given geodesic distances $r$ we craft samples transformed with random transformations and measure misclassification rate of $f$. 
\begin{equation}
r_{\tau}(f) = \min r\ \textnormal{s.t.\ } \mathds{P}(f(\mathbf{x^{(i)}_{\tau}})\neq f(\mathbf{x^{(i)}})\ |\ d_{\mathbf{x^{(i)}_{\tau}}}(e,\tau)=r) \geq 0.5,
\label{eq:random_robustness}
\end{equation}
where $0.5$ is a user defined threshold.
A robust model can maintain higher classification accuracy for images that have larger geodesic distance from the originals.

\subsection{ManiFool Augmentation}

A significant difference in our approach to the original ManiFool work is that our purpose is not to fool a deep neural network and craft an adversarial example~\cite{intriguing_properties}, but rather to utilize the transformed images for data augmentation. Therefore, once we compute the affine transformation $\tau_{i_{\max}}$ that crosses the decision boundary and fools $f$, we backtrack onto the original class manifold $\mathcal{M}$ via an iterative reduction of the final step size.
\begin{figure}[t]
  \begin{minipage}[t]{0.5\textwidth}
\centering
  \includegraphics[width=\textwidth]{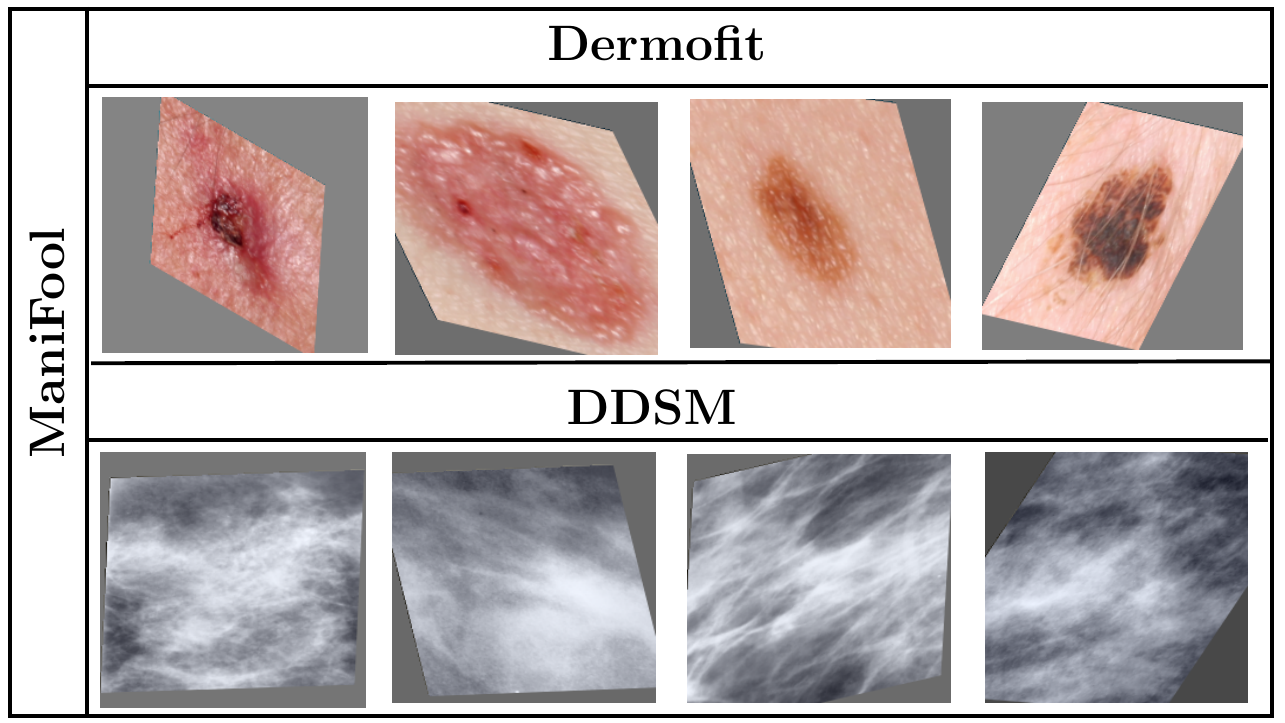}
  \end{minipage}
  \hfill
    \begin{minipage}[t]{0.45\textwidth}
    \vspace{-3cm}
  \captionof{figure}{\small{Examples generated with ManiFool Augmentation for the two datasets, namely Dermofit and DDSM.}}
  \label{fig:manifool_examples}
  \end{minipage}
  \vspace{-0.2cm}
\end{figure}
Initially, for all the images in the training set of the given dataset, we create ManiFool Augmentation samples that reside around the edges of the class manifolds with an independent black-box classifier $f$. 
Afterwards, we mix the generated samples with the original data in an equal ratio and train a model from scratch. An alternative approach would have been to utilize all the geometrically transformed images at every step $i$ towards the decision boundary for data augmentation. However, it was crucial to maintain an equal ratio of transformed and original samples in the final dataset, so that models utilizing it for training would not be biased to geometrically transformed images, due to an imbalanced amount of samples. Hence, we only utilized the transformed samples in the vicinity of the decision boundary, to provide the maximum possible variance to the models during training.
Samples crafted with ManiFool Augmentation are presented in Fig.~\ref{fig:manifool_examples}.















\section{Experimental Setup}
\paragraph{\textnormal{\textbf{Datasets}}} ManiFool Augmentation has been validated on two challenging, public, medical imaging classification datasets, namely, Digital Database for Screening Mammography (DDSM)~\cite{ddsm_1},~\cite{ddsm_2} and Dermofit~\cite{dermofit}. DDSM consists of 11.617 expert selected regions of interest (ROI) of mammograms from 1861 patients annotated as normal, benign or malignant by radiologists.  Dermofit is an image library consisting of 1300 high-quality dermatoscopic images, with histologically validated fine-grained expert annotations (10 classes). Both datasets were split at patient-level with non-overlapping folds (70\% training and 30\% testing).

\paragraph{\textnormal{\textbf{Model Training}}}
Three state-of-the-art architectures, namely ResNet18~\cite{resnet}, VGG16~\cite{VGG} and InceptionV3~\cite{inception}, were used for the evaluation.
All networks were initialized with ImageNet weights, therefore appropriate resizing and normalization of the input were performed.
The loss function selected for the aforementioned classification problems was weighted Cross Entropy, since the selected datasets are characterized by severe class imbalance.
Class weights were computed with median frequency balancing, as described in~\cite{error_corrective_boosting}.
The models were optimized with Adam optimizer with an initial learning rate of $0.001$ across the board.
The experiments were implemented in the deep learning framework PyTorch~\cite{pytorch} and an NVIDIA Titan Xp was used to train the models for $50$ epochs.

\paragraph{\textnormal{\textbf{Baseline Methods}}}
To validate the proposed contributions we perform not only ablative studies but also comparison against other widely used augmentation techniques. 
ManiFool Augmentation was compared with models trained without any augmentation (referred to as "None" in the following Section) and models trained with traditional random augmentation ("Random"), i.e. rotation and horizontal flipping.
The proposed method (noted as "ManiFool" in the tables of results) was also evaluated against augmentation techniques including Random Erasing~\cite{random_erasing} ("Erasing"), a commonly used and fast augmentation technique that replaces random patches of the image with Gaussian noise, and data augmentation with images synthesized by GANs ("DCGAN"), following the method described in~\cite{gan_isbi}. 

\paragraph{\textnormal{\textbf{ManiFool Augmentation Crafting}}}
A noteworthy implementation detail is that for the crafting of the ManiFool Augmentation samples, black-box state-of-the-art models were utilized as the differential classifier $f$ described in Section~\ref{methodology}. Those models were previously trained on the given datasets but are not utilized in the evaluation phase of this work, to avoid any bias and to ensure that the dataset is previously unseen by all the evaluated models.

\section{Results and Discussion}
In this Section the detailed results of the ablative evaluation, as well as the baseline comparisons will be discussed, along with the effects of the proposed method to the performance and robustness of the models.

\paragraph{\textnormal{\textbf{Performance improvement with ManiFool Augmentation}}}
Tables~\ref{tab:dermofit_results} and~\ref{tab:ddsm_results}
report the results of the ablative and baseline evaluation of the proposed Mani\-Fool Augmentation method for the Dermofit and DDSM Datasets. Initially, it can be observed that the performance of models without any augmentation is significantly lower, due to overfitting and limited manifold exploration.
Random Augmentation provides an improvement in performance but offers no guarantee regarding the increase in the variance that the model is exposed to during training. Moreover, random augmentation can result in out-of-distribution samples, which could hinder model training. 
Augmented samples created by ManiFool are guaranteed to originate from the same distribution as the original training data, a trait particularly crucial in the setting of medical applications, where misclassifications can have severe and undesired outcomes.
Furthermore, Manifool Augmentation, with its improved exploration capabilities, increases the accuracy by $2\%-3\%$ across both datasets and model architectures.
Additionally, ManiFool Augmentation consistently outperforms Random Erasing, Random Augmentation and GAN Augmentation by approximately $2\%$ across datasets and models.

\begin{figure}[t]
\centering
  \begin{minipage}[t]{0.65\textwidth}
\centering
\bgroup
\def\arraystretch{1.5}
\setlength\tabcolsep{8pt}
\resizebox{\textwidth}{!}{
\begin{tabular}{clc|c|c|c|}
& & \textbf{\scriptsize{None}} & \textbf{\scriptsize{Random}} & \textbf{\scriptsize{Erasing}} & \textbf{\scriptsize{ManiFool}} \\
\hline
\multicolumn{1}{c|}{\multirow{3}{*}{\rotatebox[origin=c]{90}{\scriptsize{\textbf{ResNet}}}}}    & \multicolumn{1}{l|}{\textbf{\scriptsize{Original Test}}}     & 0.7379                   & 0.7859                       & 0.7867                  & \textbf{0.8126}                \\ \cline{2-6} 
\multicolumn{1}{c|}{}                                      & \multicolumn{1}{l|}{\textbf{\scriptsize{Random Affine}}}     & 0.6515                   & 0.6962                       & 0.6573                  & \textbf{0.7900}                \\ \cline{2-6}
\multicolumn{1}{c|}{}                                      & \multicolumn{1}{l|}{\textbf{\scriptsize{Random Projectve}}}  & 0.4373                   & 0.4817                       & 0.4555                  & \textbf{0.6263}                \\ \hline
\multicolumn{1}{c|}{\multirow{3}{*}{\rotatebox[origin=c]{90}{\scriptsize{\textbf{VGG}}}}}      & \multicolumn{1}{l|}{\textbf{\scriptsize{Original Test}}}     & 0.7526                   & 0.8080                       & 0.7924                  & \textbf{0.8258}                \\ \cline{2-6} 
\multicolumn{1}{c|}{}                                      & \multicolumn{1}{l|}{\textbf{\scriptsize{Random Affine}}}     & 0.6993                   & 0.7387                       & 0.6751                  & \textbf{0.8011}                \\ \cline{2-6} 
\multicolumn{1}{c|}{}                                      & \multicolumn{1}{l|}{\textbf{\scriptsize{Random Projective}}} & 0.4319                   & 0.5140                       & 0.5071                  & \textbf{0.6200}                \\ \hline
\multicolumn{1}{c|}{\multirow{3}{*}{\rotatebox[origin=c]{90}{\scriptsize{\textbf{Inception}}}}} & \multicolumn{1}{l|}{\textbf{\scriptsize{Original Test}}}     & 0.7303                   & 0.8051                       & 0.7898                  & \textbf{0.8275}                \\ \cline{2-6} 
\multicolumn{1}{c|}{}                                      & \multicolumn{1}{l|}{\textbf{\scriptsize{Random Affine}}}     & 0.5544                   & 0.7063                       & 0.7123                  & \textbf{0.7883}                \\ \cline{2-6} 
\multicolumn{1}{c|}{}                                      & \multicolumn{1}{l|}{\textbf{\scriptsize{Random Projective}}} & 0.2149                   & 0.4388                       & 0.4630                  & \textbf{0.5376}                \\ \hline
\end{tabular}
}
\egroup
\end{minipage}
  \hfill
  \begin{minipage}[t]{0.3\textwidth}
  \vspace{-1.7cm}
\captionof{table}{Comparative evaluation of models trained on Dermofit using different augmentation techniques and ManiFool Augmentation.}
  \label{tab:dermofit_results}
    \end{minipage}
\end{figure}

\paragraph{\textnormal{\textbf{Limitations of Augmentation with GANs}}}
Generating synthetic images utilizing GANs is a task widely investigated recently as was discussed earlier in Section~\ref{related_work}.
However, limitations occur regarding GANs for medical imaging: In most cases the resolution of the synthetic images is low leading to a substantial loss of information and quality.
Furthermore, GANs trained on the entire dataset do not provide the ground truth label of the generated samples.
Therefore in order to use synthetic images for data augmentation with their respective label we have to train $n$ conditional GANs~\cite{dc_gans}, where $n$ represents the number of classes.
This is both time consuming and sometimes, unachievable due to limited data.
\begin{figure}[t]
\centering
  \begin{minipage}[t]{0.65\textwidth}
\centering
\bgroup
\def\arraystretch{1.5}
\setlength\tabcolsep{8pt}
\resizebox{\textwidth}{!}{
\begin{tabular}{clc|c|c|c|c|}
& & \textbf{\scriptsize{None}} & \textbf{\scriptsize{Random}} & \textbf{\scriptsize{Erasing}} &
\textbf{\scriptsize{DCGAN}} &
\textbf{\scriptsize{ManiFool}} \\
\hline
\multicolumn{1}{c|}{\multirow{3}{*}{\rotatebox[origin=c]{90}{\scriptsize{\textbf{ResNet}}}}}    & \multicolumn{1}{l|}{\textbf{\scriptsize{Original Test}}}     & 0.8321                   & 0.8254                       & 0.8294 & 0.8228 & \textbf{0.8426}                \\ \cline{2-7} 
\multicolumn{1}{c|}{}                                      & \multicolumn{1}{l|}{\textbf{\scriptsize{Random Affine}}}     & 0.7225                   & 0.6849                       & 0.6073 & 0.6964                  & \textbf{0.7970}                \\ \cline{2-7}
\multicolumn{1}{c|}{}                                      & \multicolumn{1}{l|}{\textbf{\scriptsize{Random Projectve}}}  & 0.2483                   & 0.2078                       & 0.3245 & 0.2657                  & \textbf{0.3245}                \\ \hline
\multicolumn{1}{c|}{\multirow{3}{*}{\rotatebox[origin=c]{90}{\scriptsize{\textbf{VGG}}}}}      & \multicolumn{1}{l|}{\textbf{\scriptsize{Original Test}}}     & 0.7914                   & 0.8381                       & 0.8377 & 0.8405                  & \textbf{0.8443}                \\ \cline{2-7} 
\multicolumn{1}{c|}{}                                      & \multicolumn{1}{l|}{\textbf{\scriptsize{Random Affine}}}     & 0.2444                   & 0.6547                       & 0.7194 & 0.7371                  & \textbf{0.8094}                \\ \cline{2-7} 
\multicolumn{1}{c|}{}                                      & \multicolumn{1}{l|}{\textbf{\scriptsize{Random Projective}}} & 0.1901                   & 0.2046                       & 0.2388 & 0.2279                  & \textbf{0.2733}                \\ \hline
\multicolumn{1}{c|}{\multirow{3}{*}{\rotatebox[origin=c]{90}{\scriptsize{\textbf{Inception}}}}} & \multicolumn{1}{l|}{\textbf{\scriptsize{Original Test}}}     & 0.8438                  & \textbf{0.8454}                       & 0.8424 & 0.8414       & 0.8451                \\ \cline{2-7} 
\multicolumn{1}{c|}{}                                      & \multicolumn{1}{l|}{\textbf{\scriptsize{Random Affine}}}     & 0.4854                   & 0.6423                       & 0.6006 & 0.6980                  & \textbf{0.7330}                \\ \cline{2-7} 
\multicolumn{1}{c|}{}                                      & \multicolumn{1}{l|}{\textbf{\scriptsize{Random Projective}}} & 0.1954                   & 0.2164                       & 0.2019 & 0.1980                  & \textbf{0.2356}                \\ \hline
\end{tabular}
}
\egroup
\end{minipage}
  \hfill
  \begin{minipage}[t]{0.3\textwidth}
  \vspace{-1.7cm}
\captionof{table}{Comparative evaluation of models trained on DDSM using different augmentation techniques and ManiFool Augmentation.}
  \label{tab:ddsm_results}
    \end{minipage}
\end{figure}
For example, some classes of the Dermofit dataset only have 23 samples for training, making training a conditional GAN on 23 images extremely challenging, if at all possible.
Attempts have been made to solve the GAN labelling problem in the medical context~\cite{gan_rueckert}, by generating Brain CT scans along with a paired segmentation label map.
However, this approach does not offer any guarantee on the correctness of the label maps and though the performance increase on the test set looks promising, mislabeling could induce ambiguity during training and jeopardize the robustness of the model.

Additionally, compared to Manifool Augmentation, augmentation with GANs does not guarantee increase in the variance to which the model is exposed, since images are sampled randomly from the training data distribution and not from the outer regions of the manifold as can be seen in Fig.~\ref{fig:overview}.



\paragraph{\textnormal{\textbf{Robustness on Random Geometric Transformations}}}
A noteworthy finding highlighted in Tables~\ref{tab:dermofit_results} and~\ref{tab:ddsm_results} is the significant increase in the robustness of models trained with ManiFool Augmentation to random transformations. The improvement is not only impressive, because it ranges from $7\%$ to $15\%$, but also because even though the proposed augmentation exclusively utilized affine transformations, the robustness to projective ones was drastically improved as well. The remaining evaluated augmentation techniques, i.e. Random Erasing and GAN augmentation, provided much lower, if any, improvement in the robustness of the networks in comparison to the standard random augmentation.

\begin{figure}[ht]
\begin{minipage}[t]{0.6\textwidth}
\centering
  \includegraphics[width=\linewidth]{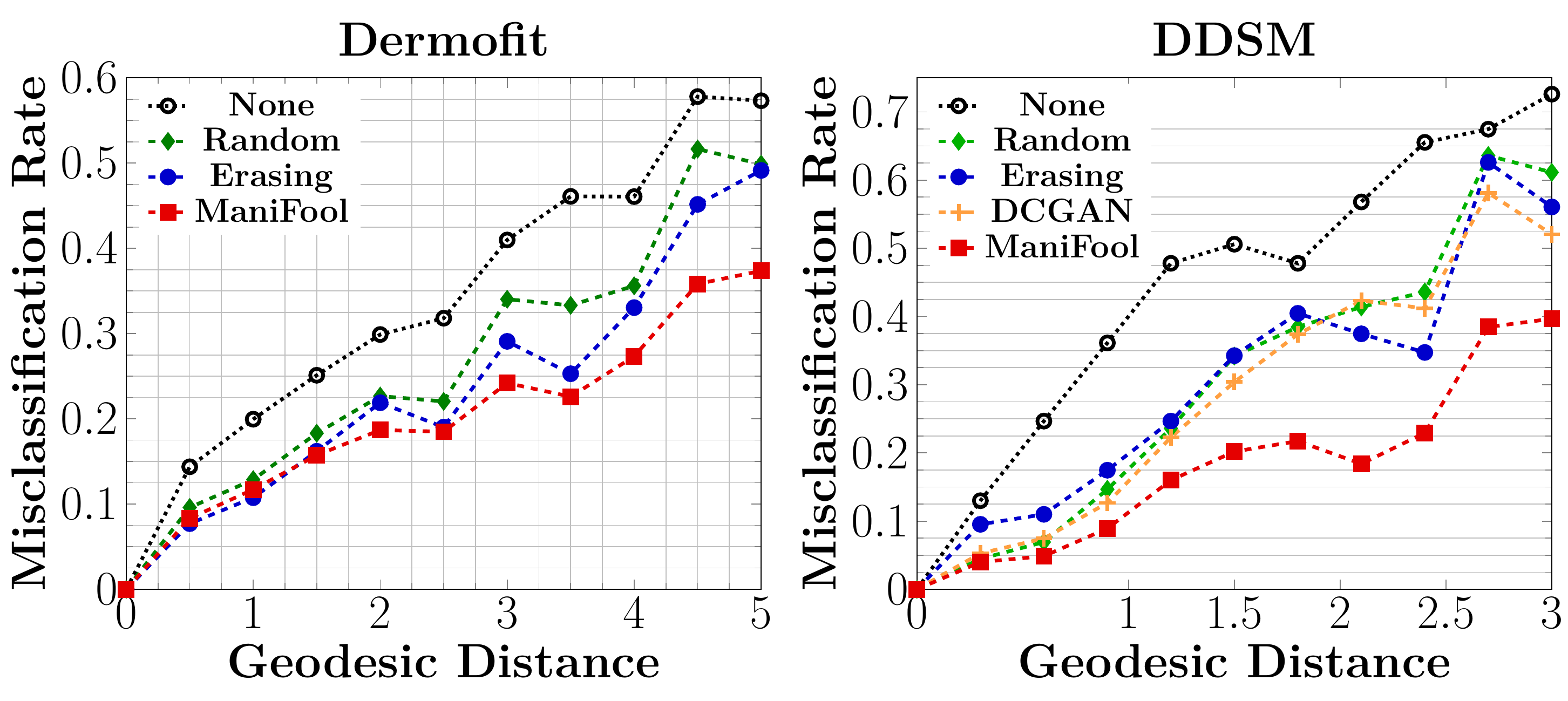}
\end{minipage}
\hfill
\begin{minipage}[t]{0.35\textwidth}
\vspace{-3cm}
  \captionof{figure}{\small{Robustness of models with different augmentation methods to random transformations with increasing geodesic distance.}}
  \label{fig:robustness_curve}
\end{minipage}
\end{figure}

\begin{figure}[t]
  \begin{minipage}[t]{0.5\textwidth}
\centering
  \includegraphics[width=\textwidth]{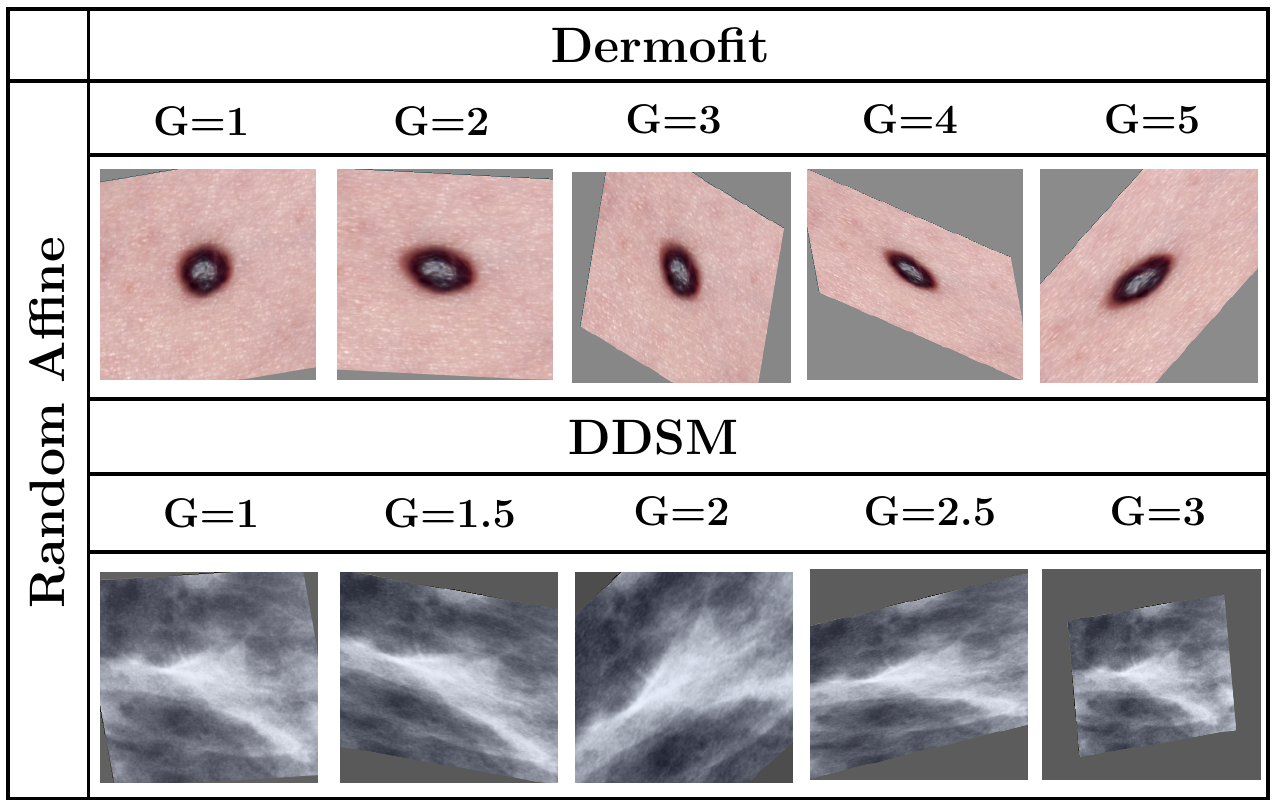}
  \end{minipage}
  \hfill
    \begin{minipage}[t]{0.4\textwidth}
    \vspace{-3.5cm}
  \captionof{figure}{\small{Examples generated with Random Affine Transformations for Dermofit~\cite{dermofit} and DDSM~\cite{ddsm_1} for a specific range of Geodesic Distances $G$.}}
  \label{fig:random_samples}
  \end{minipage}
\end{figure}

Another experiment evaluating the effect of the ManiFool Augmentation in the robustness of the trained models is shown in Fig.~\ref{fig:robustness_curve}. As described in Section~\ref{methodology}, Equation~\ref{eq:random_robustness} evaluates the misclassification rate of a classifier for samples transformed with random affine transformations for a given range of geodesic distance scores. 
In Fig.~\ref{fig:random_samples} we show images generated within a range of $G\in[1,5]$ for Dermofit and $G\in[1,3]$ that were used to evaluate the misclassification rates of the evaluated models. As can be seen in Fig.~\ref{fig:robustness_curve}, the models trained with Mani\-Fool Augmentation achieve significantly lower misclassification rates for larger values of the geodesic distance $G$.
\vspace{-0.2cm}
\paragraph{\textnormal{\textbf{Effect on Cross-Dataset Performance}}}
In order to showcase the improved robustness provided by the ManiFool Augmentation, we perform cross-dataset evaluation between Dermofit and HAM10000~\cite{ham10000}, which consists of 10.000 skin lesion images and there are 7 overlapping classes between the two datasets. Notably all models trained with the proposed method, achieve $1\%-5\%$ higher accuracy on the unseen dataset, as can be observed in Table~\ref{tab:cross_dataset}. This validates the hypothesis that ManiFool Augmentation improves the model's understanding of the underlying data distribution and leads to the increase of the model's robustness not only on geometric transformations, but also on unseen test samples.
\vspace{-0.4cm}
\begin{table}[t]
\bgroup
\def\arraystretch{1.2}
\setlength\tabcolsep{9pt}
\resizebox{\textwidth}{!}{
\begin{tabular}{l|c|c|c|c|c|c|c|c|}
                                         & \multicolumn{2}{c|}{\textbf{None}}  & \multicolumn{2}{c|}{\textbf{Random}} & \multicolumn{2}{c|}{\textbf{Erasing}} & \multicolumn{2}{c|}{\textbf{ManiFool}} \\ \cline{2-9} 
                                         & \textbf{Dermofit} & \textbf{HAM10k} & \textbf{Dermofit}  & \textbf{HAM10k} & \textbf{Dermofit}  & \textbf{HAM10k}  & \textbf{Dermofit}   & \textbf{HAM10k}  \\ \hline
\multicolumn{1}{|l|}{\textbf{ResNet}}    & 0.7379            & 0.1983          & 0.7859             & 0.3847          & 0.7867             & 0.1699           & \textbf{0.8136}     & \textbf{0.3854}  \\ \hline
\multicolumn{1}{|l|}{\textbf{VGG}}       & 0.7526            & 0.1911          & 0.8080             & 0.3101          & 0.7924             & 0.1947           & \textbf{0.8238}     & \textbf{0.3419}  \\ \hline
\multicolumn{1}{|l|}{\textbf{Inception}} & 0.7303            & 0.2798          & 0.8051             & 0.2520          & 0.7898             & 0.2140           & \textbf{0.8275}     & \textbf{0.3009}  \\ \hline
\end{tabular}
}
\egroup
\vspace{0.2cm}
\caption{Comparative evaluation of models trained on Dermofit with different augmentation methods and deployed on HAM10k, an unseen skin lesion classification dataset.}
\label{tab:cross_dataset}
\vspace{-0.5cm}
\end{table}
\vspace{0.2cm}
\begin{minipage}[t]{0.5\textwidth}
\vspace{0.1cm}
\hspace{-0.45cm}
\bgroup
\def\arraystretch{1.1}
\setlength\tabcolsep{9pt}
\resizebox{\textwidth}{!}{
\begin{tabular}{lccc}
\multirow{1}{*}{\textbf{}}              & \multicolumn{3}{c}{\textbf{Geodesic Distance}}                                                                           \\ \cline{2-4} 
                                        & \multicolumn{1}{|c|}{\textbf{ResNet}} & \multicolumn{1}{c|}{\textbf{VGG}} & \multicolumn{1}{l|}{\textbf{Inception}} \\ \hline
\multicolumn{1}{|l|}{\textbf{Dermofit}} & \multicolumn{1}{c|}{2.128}             & \multicolumn{1}{c|}{2.660}          & \multicolumn{1}{c|}{\textbf{3.391}}       \\ \hline
\multicolumn{1}{|l|}{\textbf{DDSM}}     & \multicolumn{1}{c|}{\textbf{1.510}}    & \multicolumn{1}{c|}{1.240}          & \multicolumn{1}{c|}{1.242}                \\ \hline
\end{tabular}
}
\egroup
\end{minipage}
\hfill
\begin{minipage}[t]{0.5\textwidth}
\captionof{table}{Reported average robustness measure score defined in Equation~\ref{eq:robustness_measure} for different state-of-the-art architectures.}
\label{tab:robustness_results}
\end{minipage}
\paragraph{\textnormal{\textbf{Robustness of Different Architectures}}}
After we utilize a classifier $f$ to craft ManiFool Augmentation samples, we can calculate the average geodesic distance between the original and transformed samples (Equation~\ref{eq:robustness_measure}). This measure can quantify the robustness of a machine learning model, since it implicitely measures the distance between the learned decision boundaries. Therefore, models that achieve higher robustness will be characterized by a larger geodesic distance between classes. In previous works, such as~\cite{paschali_miccai}, attempts have been made to evaluate the robustness of a classifier utilizing adversarial examples. However, such examples cannot appear naturally and no quantitative measures have been given regarding the robustness. In this work, after we generated the ManiFool Augmentation samples we calculated the robustness scores for the given classifiers, that can be seen in Table~\ref{tab:robustness_results}. This experiment showcases how the robustness of different architectures can flunctuate according to the given dataset. Therefore, it is not sufficient to utilize a state-of-the-art architecture, based on its results on an independant dataset, since its robustness can significantly vary. In our case, InceptionV3 was the most robust model for the Dermofit dataset, while ResNet18 achieved the highest robustness score for DDSM.

\section{Conclusion}
In this paper we proposed a novel data augmentation technique based on affine geometric transformations and quantified the robustness of machine learning classifiers. Experiments on challenging medical imaging tasks, namely fine grained skin lesion classification and mammogram tumor classification showcased the advantages of the proposed ManiFool Augmentation. On one hand the performance achieved by the evaluated models increased for the original test set and outperformed other commonly used data augmentation techniques. On the other hand, the robustness of the models trained with the proposed augmentation scheme was increased both for random affine and projective transformations but also cross-datasets, in an unseen test scenario. Furthermore, a qualitative measure for the robustness of machine learning classifiers was calculated and showcased the variations in the robustness of state-of-the-art models for different datasets. Future work includes extension of the ManiFool Augmentation to a wider range or transformations for a variety of medical imaging tasks. 

%
%

\end{document}